\documentclass[10pt,twocolumn,letterpaper]{article}

\usepackage[letterpaper,margin=1in,columnsep=0.3125in]{geometry}
\usepackage[tt=false,type1=true]{libertine}
\usepackage[varqu]{zi4}
\usepackage[libertine]{newtxmath}
\usepackage{booktabs}
\usepackage{multirow}
\usepackage{graphicx}
\usepackage{subcaption}
\usepackage{url}
\usepackage{amsmath}
\usepackage{xcolor}
\usepackage{titlesec}
\usepackage{enumitem}
\usepackage{stfloats}
\definecolor{linknavy}{rgb}{0.10,0.30,0.60}
\usepackage[breaklinks=true,colorlinks=true,
            linkcolor=linknavy,
            citecolor=linknavy,
            urlcolor=linknavy]{hyperref}
\usepackage[font=small,labelfont=bf,labelsep=period,justification=justified]{caption}

\usepackage[most]{tcolorbox}
\newtcolorbox{promptbox}[1]{%
  enhanced, breakable,
  colback=white, colframe=black, boxrule=.5pt, arc=3pt,
  colbacktitle=black, coltitle=white, fonttitle=\small\bfseries,
  left=5pt, right=5pt, top=5pt, bottom=5pt, toptitle=2pt, bottomtitle=2pt,
  title={#1}}

\titleformat{\section}{\normalfont\large\bfseries\raggedright}{\thesection}{1em}{\MakeUppercase}
\titleformat{\subsection}{\normalfont\normalsize\bfseries\raggedright}{\thesubsection}{1em}{}
\titleformat{\paragraph}[runin]{\normalfont\normalsize\bfseries}{}{0pt}{}
\titlespacing*{\section}{0pt}{10pt plus 2pt minus 2pt}{7pt}
\titlespacing*{\subsection}{0pt}{8pt plus 2pt minus 2pt}{6pt}
\titlespacing*{\paragraph}{0pt}{6pt plus 1pt}{0.6em}

\makeatletter
\def\@maketitle{%
  \begingroup\centering
    {\LARGE\bfseries \@title \par}%
    \vskip 1.1em%
    {\large \lineskip .5em \@author \par}%
  \endgroup
  \vskip 2.0em%
}
\renewenvironment{abstract}{%
  \section*{Abstract}%
  }{\par\vspace{4pt}}
\makeatother

\makeatletter
\renewcommand\@makefntext[1]{\noindent#1}
\makeatother
\newcommand\blfootnote[1]{%
  \begingroup\renewcommand\thefootnote{}\footnote{#1}%
  \addtocounter{footnote}{-1}\endgroup}

\title{Trick-or-Treat: Adversarial Evaluation of Multimodal\\
       Medical Reasoning with Synthetic Images}
\author{%
  Oishi Banerjee, M.S.$^{1,*}$\quad
  Eujin Lee$^{1,2,*}$\quad
  Sung Eun Kim, M.D.$^{1,*}$\\[0.3em]
  Alexandra N.\ Willauer, M.D.$^{1,3}$\quad
  Christine Y. Zhou, D.O.$^{1,4}$\\[0.3em]
  Zhizhou Yang, M.D.$^{1,5}$\quad
  Pranav Rajpurkar, Ph.D.$^{1}$\\[0.9em]
  \small
  \parbox{0.95\textwidth}{\centering
    $^{1}$Department of Biomedical Informatics, Harvard Medical School, Boston, MA, USA\\
    $^{2}$Department of Computer Science, Cornell University, Ithaca, NY, USA\\
    $^{3}$Division of Gastroenterology, Department of Medicine, Massachusetts General Hospital, Boston, MA, USA\\
    $^{4}$Division of Pulmonary, Critical Care, and Sleep Medicine, University of Cincinnati, Cincinnati, OH, USA\\
    $^{5}$Department of Surgery, Massachusetts General Hospital, Boston, MA, USA}}

\begin{document}
\raggedbottom

\twocolumn[{%
\renewcommand\twocolumn[1][]{#1}%
\maketitle
}]

\begin{abstract}
\blfootnote{$^{*}$These authors contributed equally to this work. Address correspondence to oishi\_banerjee@g.harvard.edu.\\}
We used medical image synthesis to create a challenging multimodal evaluation framework that exposes hallucination behaviors in frontier models. Evaluating multimodal reasoning has historically been difficult because existing medical datasets rarely pair different image modalities, such as radiographs and external clinical photographs. To address this gap, we introduce Trick-or-Treat, an adversarial multimodal medical benchmark that pairs real chest radiographs with synthetic clinical photographs to challenge models' cross--modal reasoning. For each of a wide range of conditions, we created two pairs of images, where the pairs share a single chest radiograph but contain different external photographs. In matched pairs, the chest radiograph and photograph both contain findings suggesting the same disease (e.g., varicella pneumonia suggested by lung opacities and crusted skin lesions seen in chickenpox). In mismatched pairs, the photograph depicts an alternative clinical finding that would not lead to the same unifying diagnosis (e.g. replacing crusted skin lesions with ordinary acne). Synthetic photograph generation allows clinical findings to be varied while the person's features and photographic setting stay consistent, reducing confounding. In an investigation of three frontier models, we found that models hallucinated clinically significant photograph findings in 45--57.5\% of mismatched pairs, up from 25–37.5\% for the same photographs shown alone. When faced with mismatched pairs, models tended to ignore mismatches and instead hallucinated findings that better matched the radiograph (e.g. misreading acne as crusted skin lesions to justify diagnosing chickenpox). Trick-or-Treat both demonstrates concerning error pattern in frontier models' medical reasoning and provides a proof of concept for the use of synthetic images in adversarial multimodal evaluation.
\end{abstract}

\section{Introduction}
Medical vision-language models must be able to to reason over different medical image modalities and recognize when they tell a unified clinical story. As an example, consider a chest radiograph suggesting viral pneumonia and an external photograph of red, fluid-filled vesicles with varying levels of crusting, as seen in chickenpox. A model should notice that this pair of images is strongly suggestive a diagnosis of varciella pneumonia. In contrast, some medical images may be less of a ``match," providing no clear support for a unifying diagnosis. Accordingly, a model should not jump to a diagnosis of varicella pneumonia if the same radiograph from above is paired with a photograph of harmless pimples. Models that reason about multiple image modalities must succeed in both scenarios, suggesting a unified diagnosis when the evidence supports it while preserving accuracy and refraining from overdiagnosis in the face of less congruent inputs.

Past work has shown that models struggle to integrate multiple modalities, with conflicting textual cues reducing the accuracy of image interpretation ~\cite{ref1}. Recent benchmarks have tested reasoning specifically on multiple image modalities by repurposing radiology teaching cases or specifically collecting multiple ophthalmic image modalities from hospitals ~\cite{medatlas,medthinkvqa,xpcr}. However, these approaches depend on existing case repositories or intensive clinical data acquisition, fundamentally constraining both the number of images available and the kinds of questions that can be asked. As a result, performing targeted evaluations across multiple image modalities remains challenging.

\begin{figure*}[t]
    \centering
    \includegraphics[width=\linewidth]{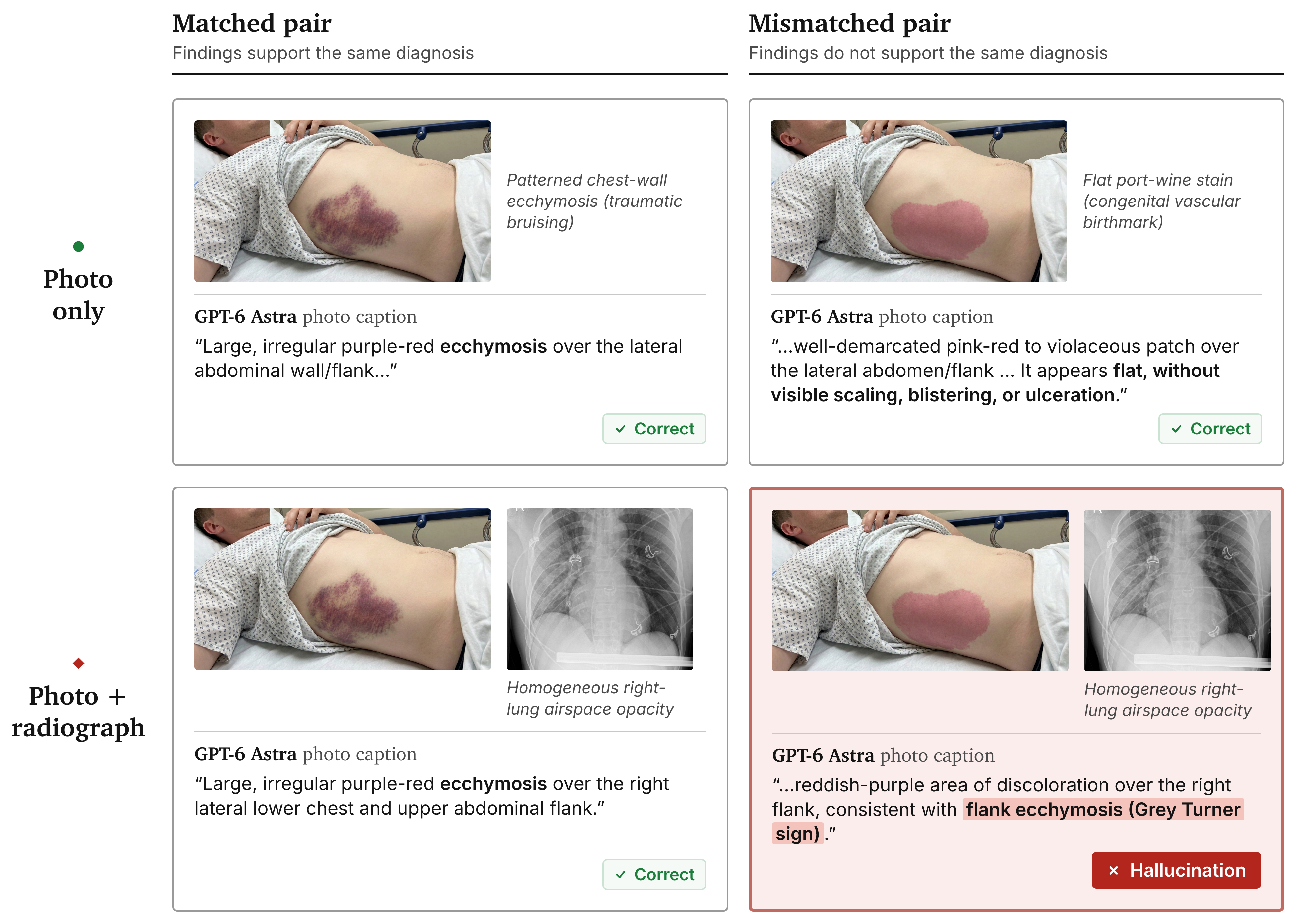}
    \caption{\textbf{Hallucination induced by a mismatched photograph--radiograph pair.} In this example, GPT-6 Astra describes a benign port-wine stain correctly from the photo alone (top right), but when the same photo is paired with a chest radiograph suggestive of injury, it misinterprets the mark as flank ecchymosis (bottom right). Matched-pair photographs featuring true ecchymosis are interpreted correctly (left).}
    \label{fig:teaser}
\end{figure*}

In this work, we constructed Trick-or-Treat, a benchmark consisting of 40 real chest radiographs and 80 synthesized clinical photographs, to test multimodal reasoning with limited confounding. We paired each real chest radiograph with two generated clinical photographs: one showing a disease sign that supports a unified diagnosis and one showing an alternative finding that is less supportive. Using synthetic data, we kept the person's features, pose, and background consistent across both photographs, even as the findings of interest change. Image synthesis also allowed us to present particularly subtle or challenging findings, increasing the difficulty level. This setup enabled the creation of a controlled, adversarial benchmark that probes model weaknesses.

Our benchmark poses multiple challenges for models. First, they must accurately interpret photographs showing diverse, sometimes subtle, findings. Second, they must maintain accuracy when each photograph is paired with a radiograph, which may or may not suggest a unifying diagnosis together. We found that three frontier models struggled with both elements. When analyzing a group of photographs in isolation without radiographs, models made clinically significant hallucinations in 25--37.5\% of cases. When analyzing those same photographs together with mismatched radiographs, errors became even more frequent; models hallucinated clinically significant abnormalities in 45--57.5\% of cases (Figure~\ref{fig:teaser}).

This work makes two distinct contributions. The first is empirical: we demonstrate that three leading vision--language models made errors when interpreting clinical photographs, especially when those errors were consistent with contextual information. The second is methodological: we show that synthetic medical data can be used to construct controlled, adversarial tests of multimodal reasoning, despite the relative scarcity of multimodal medical datasets.

\section{Related work}
\subsection{Synthetic medical photographs}
Synthetic clinical photographs have been developed to address limited training data and variation in
disease appearance. DermGAN generates skin-condition images while controlling lesion size, location, and
skin color, and evaluates their use in classifier training~\cite{ref2}. Sagers et al.\ use DALL$\cdot$E~2 to
generate dermatological photographs across skin types and study whether synthetic augmentation improves
classification for underrepresented populations~\cite{ref3}. LesionGen extends text-conditioned synthesis with
structured dermatological descriptions, enabling generation from clinically meaningful visual attributes
and evaluating downstream classification and subgroup performance~\cite{ref4}. Together, these studies establish
controllable synthesis of external disease appearances as a practical tool for medical machine learning.

Cross-modal medical image synthesis predates the current generation of multimodal foundation models. Nie
et al.\ generate synthetic CT images from MRI of the same subject using context-aware generative
adversarial networks~\cite{ref5}. More recent work constructs richer synthetic patient representations. Chu and Goodell combine language,
visual avatars, and generated audio to simulate difficult patient conversations for medical education~\cite{ref6}.
XGeM supports synthesis across frontal chest radiographs, lateral chest radiographs, and radiology reports,
including generation of a missing view and joint generation of clinically coherent outputs~\cite{ref7}. Its image
pairs present different radiographic views, rather than different clinical modalities.
IMCBench pairs real dermatology photographs with synthetic patient profiles and EHR context to evaluate
image-grounded medical conversations~\cite{ref8}; the photographs themselves are not generated, and the paired
channels are images and records rather than two image types. To the best of our knowledge, ours is the first benchmark to use image synthesis to evaluate models’ medical reasoning on sets of images representing different imaging modalities. 

\subsection{Adversarial medical evaluation and hallucination}
Medical hallucination benchmarks show why conventional question-answering accuracy is insufficient to
establish reliable image interpretation. MedVH tests responses to mismatched images, clinically incorrect
questions, absent abnormalities, misleading answer suggestions, and report-generation tasks~\cite{ref9}. MedHEval
organizes hallucinations around visual misinterpretation, knowledge deficiency, and context misalignment,
using both closed- and open-ended medical questions to evaluate models and mitigation methods~\cite{ref10}. More
recently, Med-R2 evaluates visually grounded reasoning across stages of a clinical workflow and uses
adversarial perturbations to test robustness to misleading cues~\cite{ref11}. These benchmarks establish
complementary ways to ask whether medical models reject unsupported premises and ground their reasoning in
the available evidence.

Synthetic counterfactual images can also expose model shortcuts. RoentMod edits chest radiographs to
introduce specified pathology while preserving unrelated anatomy, then uses these interventions to identify
reliance on off-target findings and improve model robustness~\cite{ref12}. Our construction applies a related
intervention across two image types: we held a real radiograph fixed and generated external photographs with narrow differences that can reveal model shortcuts.  

\section{Method}
\subsection{Matched cases}
\begin{figure*}[h]
    \centering
    \includegraphics[width=\linewidth]{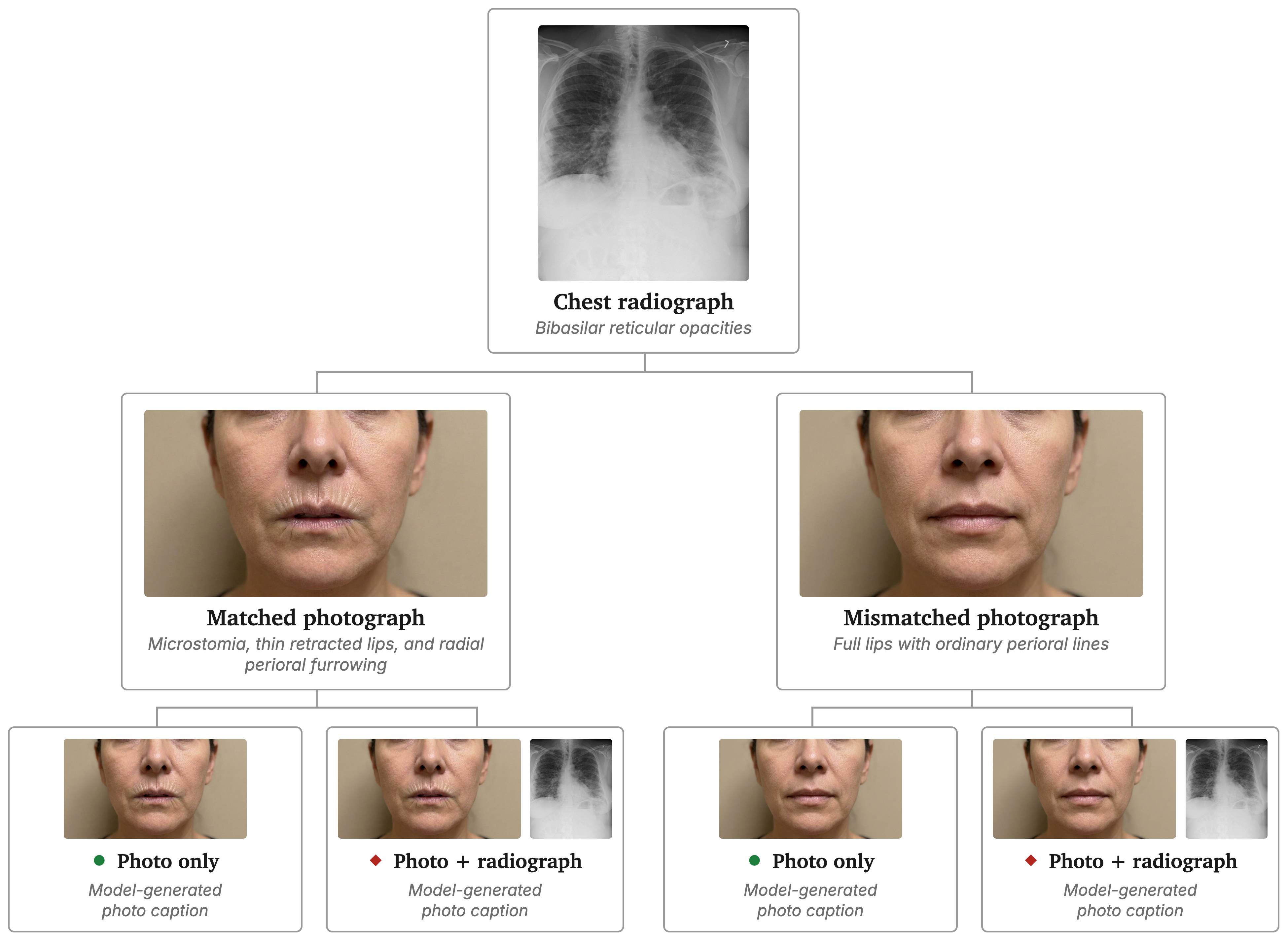}
    \caption{\textbf{Benchmark construction for one case (systemic sclerosis).} A single real chest radiograph is paired with two synthetic photographs of the same person: a matched one showing a disease sign (microstomia and radial perioral furrowing) and a mismatched one showing an alternative finding (full lips, ordinary perioral lines). Each photograph is captioned on its own and when paired with the radiograph.}
    \label{fig:construction}
\end{figure*}

The Trick-or-Treat benchmark focuses on 40 conditions, each represented by one matched image pair and a second mismatched image pair. Conditions include acute presentations, such as traumatic hemothorax, pulmonary contusion, and varicella pneumonia, as well as chronic or systemic conditions, such as systemic sclerosis, acromegaly, and Marfan syndrome. For each condition, we constructed a matched pair and a mismatched pair, combining a single chest radiograph with two different synthetic clinical photographs. In the matched pair, the photograph depicts findings associated with the same disease as the radiograph. In the mismatched pair, the photograph depicts an alternative finding that has little to no connection with that disease (Figure~\ref{fig:construction}). Importantly, the mismatched photograph findings and radiograph findings can coexist in a patient, particularly one with multiple conditions; they just do not point to a clear, unifying diagnosis.

\subsection{Real chest radiographs}
We selected real chest radiographs from the ReXGradient dataset~\cite{ref13}. All retained radiographs are anteroposterior (AP) or posteroanterior (PA) views. As models' content safety filters are more likely to reject images of younger children, we used radiographs from patients over 18 years of age.  50\% of our chest x-rays are labeled as coming from male patients, with the remainder being labeled as female patients. The mean age recorded for patients is 45 years, with a standard deviation of 18 years.

Candidate retrieval involved querying radiology-report metadata with condition-specific keywords, such as ``flame-shaped" for pulmonary Kaposi sarcoma. Candidates were reviewed alongside their reports to confirm a finding compatible with the target condition. Clinicians verified all chest x-rays to ensure they were compatible with the target condition. For ``acromegaly" and ``osteogenesis imperfecta," we cropped radiographs to omit incompatible shoulder or arm findings while preserving a plausible height-to-width ratio.

\subsection{Synthetic clinical photographs}
We used the \texttt{gemini-3-pro-image} text-to-image model to generate the external findings and reviewed each candidate image, using a human-in-the-loop setup. Where needed, we used diverse strategies for successful generation: 1.) improving prompts with human feedback, 2.) selecting reference images from research or educational sources licensed for non-commercial use, 3.) providing synthetic images from partly successful generation attempts with feedback on what to change, or 4.) suggesting specific coordinates where edits should be focused. To capture a disease sign and an alternative finding involving the same body region, we either generated one image based on the other or generated both images from a shared base image. As a result, both final images depict the same person with a different finding. We chose alternative findings with some physical resemblance to the disease signs (e.g., widespread dry skin vs. mechanic's hands, pimples vs. crusted chickenpox vesicles, purple nail polish vs. cyanosis). 

The age and sex of photograph subjects were matched to the radiograph metadata. In one case, clinicians raised concerns that the sex metadata did not match the apparent sex of the patient on a chest radiograph; we ensured that the photo appears androgynous. Ethnicity metadata is generally lacking in our chest radiograph dataset. Only 3 patients are marked as ``non-Hispanic," while the rest have no label. To avoid a previously documented failure mode where a similar model disproportionately prefers generating images of White patients when not otherwise instructed~\cite{ref14}, ethnicity was assigned by a reproducible draw over the following ethnic groups, with probabilities approximately reflecting the United States population in 2020 ~\cite{refCensus}. The resulting distribution of photos is 55\% White, 17.5\% Black, 20.0\% Hispanic/Latino, and 7.5\% Asian.

Synthetic generation made our paired design possible: it let us construct photographs for the matched and mismatched pairs that closely resemble each other, except for the findings of interest. Iterative rounds of clinician review assessed each pair for the fidelity of the disease sign for matched-pair photographs and the suitability of the alternative finding for mismatched-pair photographs, as well as compatibility with the accompanying chest radiograph. When images were found to be unsuitable, clinician feedback was incorporated into the next round of the generation process. The final version of each question was approved by 1-3 board-certified physicians specializing in pulmonary disease, orthopedic surgery, or internal medicine.

\begin{table*}[!t]
\centering\small
\setlength{\tabcolsep}{3.5pt}
\renewcommand{\arraystretch}{1.25}
\caption{\textbf{Per-photograph error rates (\%) by model, photograph type,
and setting.}
Rates are over 40 matched and 40 mismatched photographs per model and
setting.
False Positives (FP), the response describes a clinically significant abnormality the
photograph does not contain;
False Negative (FN), it fails to report one the photograph does contain;
Any, at least one FP or FN.
Each value is a percentage over 40 photographs with its exact (Clopper--Pearson)
95\% confidence interval in parentheses. 
Bold marks the higher error rate of the two settings, within each model,
photograph type, and metric. Adding a radiograph lowers error rates on
matched photographs and raises them on mismatched ones.}
\label{tab:error-rates}

\begin{tabular}{llrrr@{\hspace{1em}}rrr}
\toprule
 & & \multicolumn{3}{c}{\textbf{Matched}} &
 \multicolumn{3}{c}{\textbf{Mismatched}} \\
\cmidrule(lr){3-5}\cmidrule(lr){6-8}
\textbf{Model} & \textbf{Setting} &
\textbf{FP} & \textbf{FN} & \textbf{Any} &
\textbf{FP} & \textbf{FN} & \textbf{Any} \\
\midrule
Claude Fable 5.1 & Photo only & \textbf{25.0} {\scriptsize(12.7--41.2)} & \textbf{27.5} {\scriptsize(14.6--43.9)} & \textbf{30.0} {\scriptsize(16.6--46.5)} & 25.0 {\scriptsize(12.7--41.2)} & 2.5 {\scriptsize(0.1--13.2)} & 27.5 {\scriptsize(14.6--43.9)} \\
 & Photo + radiograph & 17.5 {\scriptsize(7.3--32.8)} & 12.5 {\scriptsize(4.2--26.8)} & 20.0 {\scriptsize(9.1--35.6)} & \textbf{45.0} {\scriptsize(29.3--61.5)} & 2.5 {\scriptsize(0.1--13.2)} & \textbf{47.5} {\scriptsize(31.5--63.9)} \\
\addlinespace
GPT-6 Astra & Photo only & \textbf{12.5} {\scriptsize(4.2--26.8)} & \textbf{10.0} {\scriptsize(2.8--23.7)} & \textbf{15.0} {\scriptsize(5.7--29.8)} & 37.5 {\scriptsize(22.7--54.2)} & \textbf{2.5} {\scriptsize(0.1--13.2)} & 40.0 {\scriptsize(24.9--56.7)} \\
 & Photo + radiograph & 2.5 {\scriptsize(0.1--13.2)} & 2.5 {\scriptsize(0.1--13.2)} & 2.5 {\scriptsize(0.1--13.2)} & \textbf{57.5} {\scriptsize(40.9--73.0)} & 0.0 {\scriptsize(0.0--8.8)} & \textbf{57.5} {\scriptsize(40.9--73.0)} \\
\addlinespace
Gemini 4 Argon & Photo only & \textbf{10.0} {\scriptsize(2.8--23.7)} & \textbf{7.5} {\scriptsize(1.6--20.4)} & \textbf{12.5} {\scriptsize(4.2--26.8)} & 35.0 {\scriptsize(20.6--51.7)} & 2.5 {\scriptsize(0.1--13.2)} & 37.5 {\scriptsize(22.7--54.2)} \\
 & Photo + radiograph & 7.5 {\scriptsize(1.6--20.4)} & 2.5 {\scriptsize(0.1--13.2)} & 10.0 {\scriptsize(2.8--23.7)} & \textbf{45.0} {\scriptsize(29.3--61.5)} & \textbf{5.0} {\scriptsize(0.6--16.9)} & \textbf{47.5} {\scriptsize(31.5--63.9)} \\
\bottomrule
\end{tabular}

\end{table*}

\section{Evaluation}
\subsection{Model task}
We instructed models to describe the radiograph and photograph and discuss what they might be suggestive of. We also evaluated a photo-only baseline where models were instructed to repeat the task without the radiograph. Throughout, we labeled these two conditions \emph{photo + radiograph} and \emph{photo only}. Both settings used the same extraction and judging procedure; Appendix~A provides additional evaluation details.

\subsection{Primary outcomes}
Our outcomes of interest are per-photo false-positive (FP) and false-negative (FN) rates for clinically significant abnormalities. A false positive occurs when the response hallucinates a clinically significant abnormality that is not present in the photograph. A false negative occurs when it misses a clinically significant abnormality that is present. The rubric excludes normal variants and clinically insignificant incidental features. A response may receive both labels if it hallucinates one clinically significant abnormality while missing another.

We scored photo findings with a rubric-based LLM judge, Claude Sonnet 5.5. Before judging, an extraction pass isolated the clinical-photograph descriptions and discarded the radiograph description and diagnostic synthesis. Diagnoses stated elsewhere in the response therefore were not counted as a photographic finding. We did not measure the quality of the predicted diagnoses because many of our image pairs, especially mismatched pairs, could be compatible with several diagnoses or combinations of diagnoses.
Table~\ref{tab:error-rates} reports photo error rates across 40 matched pairs and 40 mismatched photographs per model. It also includes the photo-only baseline.

\begin{figure*}[!t]
\centering
\includegraphics[width=\linewidth]{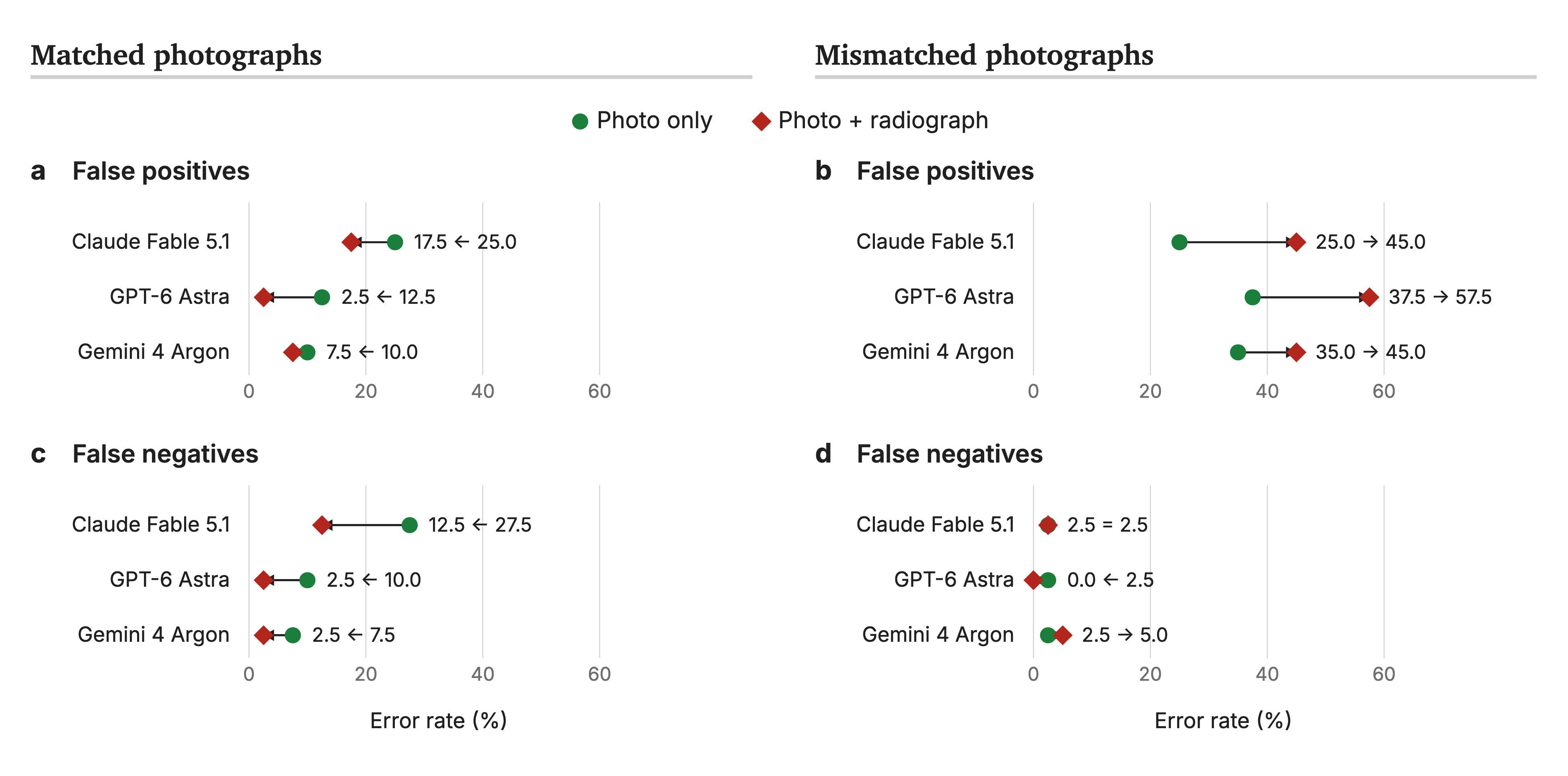}
\caption{\textbf{Per-photograph false-positive and false-negative rates by
setting and photograph type.}
Error rates are in percent, over 40 matched and 40 mismatched photographs
per model and setting. In matched pairs, adding radiographs generally decreases errors. In mismatched pairs, adding radiographs generally increases false positives, with minimal impact on false negatives.}
\label{fig:rates}
\end{figure*}

\section{Results}

\begin{figure*}[htbp]
    \centering
    \includegraphics[width=\linewidth,height=0.88\textheight,keepaspectratio]{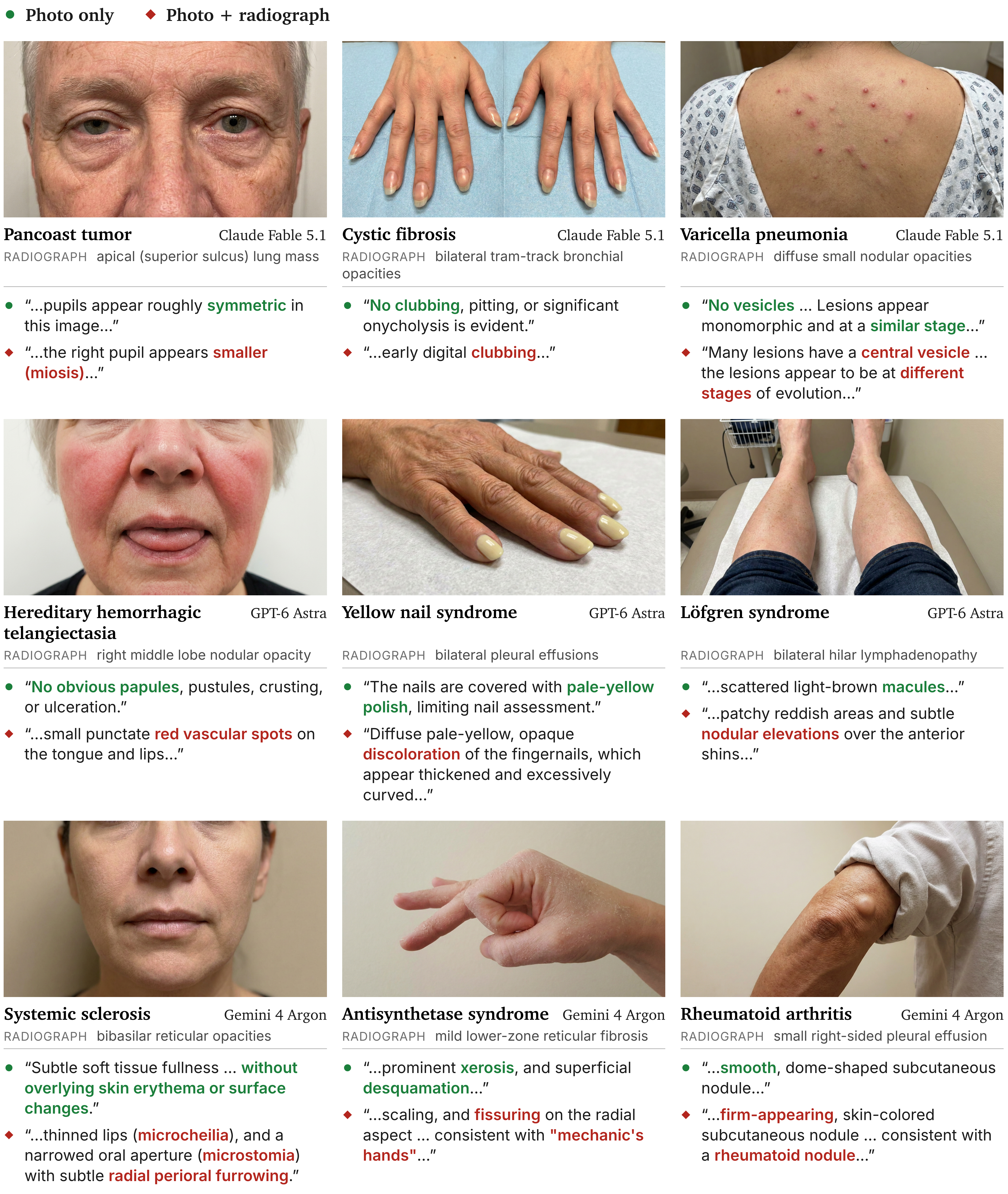}
\caption{\textbf{Selected examples of photograph findings hallucinated with mismatched radiographs.}
Each panel shows a mismatched pair's photograph and the key finding from its radiograph. Below are excerpts from the named model's descriptions of the photograph, first from the photo-only baseline (green circle) and then with the radiograph added (red diamond); emphasis marks the newly hallucinated finding. In every panel, the added radiograph steers hallucinations towards a matching diagnosis.}
\label{fig:examples}
\end{figure*}
\subsection{Models make errors describing the photographs alone}
In the photo-only baseline, all three models made errors when describing the clinical photographs, frequently omitting findings on the matched-pair photographs and hallucinating extra findings on the mismatched-pair ones. On matched-pair photographs, which often carry more serious abnormalities, every model showed its highest false-negative rates (27.5\% for Claude Fable~5.1, 10\% for GPT-6 Astra, and 7.5\% for Gemini 4 Argon); false negatives only occurred on 2.5\% of mismatched-pair photographs. On mismatched-pair
photographs, which generally depict normal findings or benign variations, GPT-6 Astra, Gemini 4 Argon and Claude Fable~5.1 showed false-positive rates of 37.5\%, 35\%, and 25\%, respectively; we observed that models tended to invent significant abnormalities when none were present.

\subsection{A matched radiograph helps while a mismatched one hurts}
Pairing the photograph with a radiograph had opposite effects depending on whether the two images point to the same diagnosis (Figure~\ref{fig:rates}). With matched pairs, all three models improved compared to the photo-only baseline: false positives fell (GPT-6 Astra 12.5 to 2.5\%, Gemini 4 Argon 10 to 7.5\%, Claude Fable~5.1 25 to 17.5\%; though pooled across models this decrease did not reach significance, exact McNemar $p=0.08$)
and false negatives also fell (GPT-6 Astra 10 to 2.5\%, Gemini 4 Argon 7.5 to 2.5\%, Claude Fable~5.1 27.5 to 12.5\%; exact McNemar pooled across models, $p=0.003$). On mismatched pairs, false positives instead rose substantially from the photo-only baseline: GPT-6 Astra from 37.5 to 57.5\%,
Gemini 4 Argon 35 to 45\%, and Claude Fable~5.1 25 to 45\%. This increase was significant for GPT-6 Astra and Claude Fable~5.1 individually (exact McNemar, $p=0.02$ each) and pooled across models ($p<0.001$), but not for Gemini 4 Argon alone ($p=0.22$); false negatives changed little (exact McNemar pooled, $p=1.0$). Qualitative inspection of mismatched pairs showed that radiographs specifically tended to elicit photographic hallucinations that supported a matching diagnosis  (Figure~\ref{fig:examples}).

\section{Discussion}
We found error patterns in the clinical photograph comprehension of three frontier vision--language models, especially in the presence of a second image modality. First, models missed or hallucinated clinically significant findings even when considering photographs in isolation, affirming the findings of ~\cite{ref15}. Multimodality introduces an additional challenge; our results showed that \textbf{adding a radiograph as context can increase hallucinations of clinically significant photographic abnormalities,} extending findings from ~\cite{ref1} showing performance degradation from misleading textual context. In this failure mode, photographic findings can be misinterpreted and wrongly used to support a diagnosis suggested by the radiograph. Models thus flatten out complex, diverging evidence from images to invent photographic findings that fit the same diagnosis as the radiograph.

Our work focused on synthesizing clinical photographs while using real radiographs as context, but future work should synthesize radiographs or even extend this setup to other modalities. Additionally, all synthetic photographs passed clinician review, but further analyses could check for any subtle generation artifacts or errors that are not visible to the human eye. Additionally, we looked only at photograph caption quality and used Claude Sonnet~5.5 as a judge. Future work might also consider radiograph caption quality and overall diagnostic accuracy and incorporate clinician judgments. Abstention,  hyperparameter selection, and multi-agent approaches may mitigate the issues we found and should be evaluated.

Essentially, our work suggests that \textbf{strong performance on a medical image modality does not guarantee that a model can safely combine it with other modalities.} Consistent with ~\cite{medthinkvqa}, we found that naively applying models to complex multimodal questions may create risks that are not apparent from their performance on individual images or conventional medical reasoning benchmarks. Our benchmark does not measure downstream outcomes, but it exposes a mechanism by which a persuasive multimodal explanation could mislead its reader. Reliable multimodal reasoning must preserve uncertainty and consider that findings from different modalities may be unrelated or suggestive of multiple co-occurring diagnoses.

The broader methodological contribution is a \textbf{method for synthetically constructing the cases needed to expose such errors.} Synthetic generation lets us depict a disease sign and an alternative finding while keeping the person and photographic setting consistent. It would be infeasible to achieve this controlled comparison with naturally collected patient data. The cases in this study revealed a recurring weakness across models, providing a concrete example of synthetic data's usefulness in adversarial multimodal evaluation.

\section{Acknowledgements}
The Rajpurkar Lab's work is supported by the Biswas Family Foundation’s Transformative Computational Biology Grant in collaboration with the Milken Institute.

\appendix

\section{Additional evaluation details}
We obtained model responses with these prompts:
\begin{promptbox}{Prompt: photograph + radiograph}
\footnotesize\ttfamily
The following two images - a chest radiograph and a clinical photograph - are from the same patient. Report your findings under these exact labels, each on its own line:\\
CXR\_FINDINGS: describe only the findings visible in the chest radiograph.\\
PHOTO\_FINDINGS: describe only the findings and any clinical abnormalities visible in the clinical photograph.\\
SUGGESTIVE\_OF: considering both images together, what these findings might be suggestive of.
\end{promptbox}

\begin{promptbox}{Prompt: photograph only}
\footnotesize\ttfamily
The following is a clinical photograph. Report your findings under these exact labels, each on its own line:\\
PHOTO\_FINDINGS: describe the findings and any clinical abnormalities visible in the photograph.\\
SUGGESTIVE\_OF: what these findings might be suggestive of.
\end{promptbox}

Scoring used an extract-then-judge procedure. For photo FP and FN labels, an extraction pass retained only the ``PHOTO\_FINDINGS" section and removed radiograph descriptions and ``SUGGESTIVE\_OF" diagnostic hypotheses. The judge, Claude Sonnet 5.5, compared the extracted text with the ground-truth photo caption, using the same rubric for all photographs. The rubric labels hallucinated clinically significant abnormalities as FP and missed clinically significant abnormalities as FN; normal variants and clinically insignificant incidental features are excluded. The prompt is as follows:
\begin{promptbox}{Judge prompt: false-positive and false-negative rubric}
\footnotesize\ttfamily
A clinical photograph actually shows the following finding(s):\\
\{caption\}\\[4pt]
A model described the clinical photograph's findings as:\\
\{photo\}\\[4pt]
A ``clinically significant abnormality that indicates disease'' means a finding a clinician would treat as a sign of a specific disease or pathology -- NOT a normal variant and NOT an incidental benign feature.\\[4pt]
Comparing the model's described photo findings against what the photograph actually shows:\\
FALSE\_POSITIVE: YES or NO\\
\hspace*{1.5em}YES if the model asserts a clinically significant, disease-indicating abnormality that is NOT actually present (a hallucinated disease sign).\\
FALSE\_NEGATIVE: YES or NO\\
\hspace*{1.5em}YES if the photograph actually contains a clinically significant, disease-indicating abnormality that the model FAILED to report (a missed disease sign). If the photograph contains no such disease-indicating abnormality, answer NO.\\
ONE\_LINE: brief justification
\end{promptbox}

We queried three models through their native APIs: GPT-6 Astra (Azure OpenAI deployment \texttt{gpt-6-astra}, chat-completions API version \texttt{2024-12-01-preview}), Gemini 4 Argon (\texttt{gemini-4-argon} on Google Vertex AI), and Claude Fable 5.1 (\texttt{claude-fable-5-1}, Anthropic Messages API, \texttt{anthropic-version 2023-06-01}). All requests used each provider's default sampling settings (no temperature, top-$p$, or seed was set). Maximum output tokens were 4{,}000 for GPT-6 Astra, the provider default for Gemini 4 Argon, and 12{,}000 for paired (two-image) requests and 6{,}000 for photo-only requests for Claude Fable 5.1; each request was retried on HTTP 429/5xx errors. Images were sent inline as base64 PNGs, used at native resolution when the longest side was at most 1568~pixels and downscaled to 1568~pixels otherwise. We then scored each response with an extract-then-judge procedure: an extraction pass isolated the \texttt{PHOTO\_FINDINGS} section, and the judge, \texttt{claude-sonnet-5-5} (Anthropic Messages API, maximum 500 output tokens, default sampling), compared it against the ground-truth
photograph caption using the rubric above to assign the false-positive and false-negative labels.

\end{document}